\documentclass{article}

\usepackage{spconf}

\usepackage{graphicx}
\usepackage{cite}
\usepackage[cmex10]{amsmath}
\usepackage{amssymb}
\usepackage{xspace}
\usepackage{xspace}
\usepackage{booktabs}
\usepackage{amsthm}
\usepackage{algorithmic}
\usepackage{enumerate}
\usepackage{balance}
\usepackage{color}
\usepackage{algorithmic}
\usepackage{algorithm}
\usepackage{mathtools}

\graphicspath{{./}{./figures/}}

\def\Vec#1{{\boldsymbol{#1}}}
\def\Mat#1{{\boldsymbol{#1}}}

\newcommand{\tr}{\mbox{tr}}

\title
  {
  Multi-Shot Person Re-Identification via Relational Stein Divergence
  }

\name
  {
  Azadeh Alavi, Yan Yang, Mehrtash Harandi, Conrad Sanderson
  }

\address
  {
  {NICTA, GPO Box 2434, Brisbane, QLD 4001, Australia}\\
  {University of Queensland, School of ITEE, QLD 4072, Australia}\\
  {Queensland University of Technology, Brisbane, QLD 4000, Australia}
  }

\begin{document}
\ninept
\maketitle

\begin{abstract}

\noindent
Person re-identification is particularly challenging due to significant
appearance changes across separate camera views. 
In order to re-identify people, a representative human signature should effectively handle differences in illumination, pose and camera parameters.
While general appearance-based methods are modelled in Euclidean spaces,
it has been argued that some applications in image and video analysis are better modelled via non-Euclidean manifold geometry.
To this end, recent approaches represent images as covariance matrices,
and interpret such matrices as points on Riemannian manifolds.
As direct classification on such manifolds can be difficult,
in this paper we propose to represent each manifold point as a vector of similarities to class representers,
via a recently introduced form of Bregman matrix divergence known as the Stein divergence.
This is followed by using a discriminative mapping of similarity vectors for final classification. 
The use of similarity vectors is in contrast to the traditional approach of embedding manifolds into tangent spaces,
which can suffer from representing the manifold structure inaccurately.
Comparative evaluations on benchmark ETHZ and iLIDS datasets for the person re-identification task
show that the proposed approach obtains better performance than recent techniques such as
Histogram Plus Epitome, Partial Least Squares,
and Symmetry-Driven Accumulation of Local Features.

\end{abstract}


\section{Introduction}
\label{sec:introduction}

\begin{figure}[!t]
  \centering
  \includegraphics[width=0.85\columnwidth,height=0.5\columnwidth]{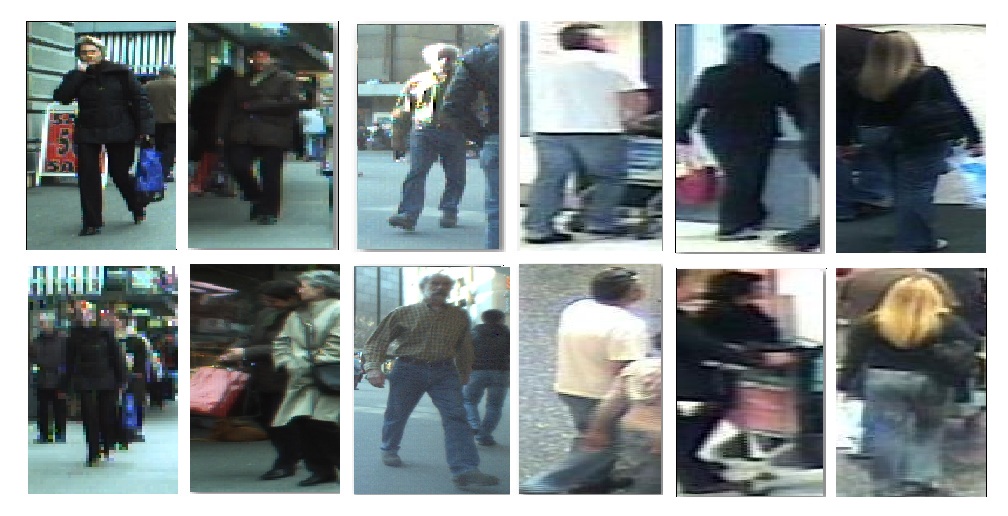}
  \vspace{-3ex}
  \caption
    {
    \small
    Examples of challenges in person re-identification,
    where each column contains images of the same person from two separate camera views.
    Challenges include pose changes, occlusions of body parts, low resolution and illumination variations. 
    }
  \label{fig:pri_examples}
\end{figure}

Person re-identification is the process of matching persons across non-overlapping camera views in diverse locations.
Within the context of surveillance, re-identification needs to function with a large set of candidates and be robust to 
pose changes, occlusions of body parts, low resolution and illumination variations.
The issues can be compounded, making a person difficult to recognise even by human observers (see Fig. \ref{fig:pri_examples} for examples).
Compared to classical biometric cues (eg.~face, gait) which may not be reliable due to non-frontality, low resolution and/or low frame-rate,
person re-identification approaches typically use the entire body.

While appearance based person re-identification methods are generally modelled in
Euclidean spaces~\cite{gray2008viewpoint,schwartz2009learning,SDALF_CVPR2010},
it has been argued that some applications in image and video
analysis are better modelled on non-Euclidean manifold geometry~\cite{turaga2011statistical}.
To this end, recent approaches represent images as covariance matrices~\cite{DPM_CVPR_2011},
and interpret such matrices as points on Riemannian manifolds~\cite{harandikernel,turaga2011statistical}.
A popular way of analysing manifolds is to embed them into tangent spaces, which are Euclidean spaces.
This process which can be interpreted as warping the feature space~\cite{turaga2008statistical}.
Embedding manifolds is not without problems,
as pairwise distances between arbitrary points on a tangent space
may not represent the structure of the manifold accurately~\cite{harandikernel,Harandi_ECCV_2012}.

In this paper we present a multi-shot appearance
based person re-identification method on Riemannian manifolds,
where embedding the manifolds into tangent spaces is not required.
We adapt a recently proposed technique for analysing Riemannian manifolds,
where points on the manifolds are represented through their similarity vectors~\cite{Alavi_WACV_2013}.
The similarity vectors contain similarities to class representers.
We obtain each similarity with the aid of a recently introduced form of Bregman matrix divergence
known as the Stein divergence~\cite{Harandi_ECCV_2012,Sra_JMLR_2012}.
The classification task on manifolds is hence converted into a task in the space of similarity vectors, 
which can be tackled using learning methods devised for Euclidean spaces,
such as Linear Discriminant Analysis~\cite{Bishop_2006}.
Unlike previous person re-identification methods,
the proposed method does not require separate settings for new datasets.

We continue the paper as follows.
In Section~\ref{sec:Background} several recent methods for person re-identification are briefly described.
The proposed approach is detailed in Section~\ref{sec:Ourapproach}.
A comparative performance evaluation on two public datasets is given In Section~\ref{sec:Experiments_RDC}.
The main findings are summarised in Section~\ref{sec:conclusions}.

~
\section{Previous Work}
\label{sec:Background}

Given an image of an individual to be re-identified, the task of person
re-identification can be categorised into two main classes.
{\bf (i)}
Single-vs-Single (SvS),
where there is only one image of each person in the gallery and one in the probe;
this can be seen as a one-to-one comparison.
{\bf (ii)}
Multiple-vs-Single (MvS), or multi-shot,
where there are multiple images of each person available in gallery and one image in the probe.
Below we summarise several person re-identification methods:
Partial Least Squares~(PLS)~\cite{schwartz2009learning},
Context based method~\cite{iLIDS_pub},
Histogram Plus Epitome (HPE)~\cite{HPE_ICPR2010},
and Symmetry-Driven Accumulation of Local Features (SDALF)~\cite{SDALF_CVPR2010}.

The PLS method \cite{schwartz2009learning}
first decomposes a given image into overlapping blocks, and extracts a rich set of features from each block.
Three types of features are considered:
textures, edges, and colours. 
The dimensionality of the feature space is then reduced by employing Partial Least Squares regression (PLSR)~\cite{PLS1956},
which models relations between sets of observed variables by means of latent variables.
To learn a PLSR discriminatory model for each person, one-against-all scheme is used~\cite{geladi1986partial}.
Nearest neighbour is then employed for classification.

The Context-based method \cite{iLIDS_pub}
enriches the description of a person by contextual visual knowledge from surrounding people.
The method represents a group by considering two descriptors:
{\bf (a)}~`center rectangular ring ratio-occurrence' descriptor,
which describes the information ratio of visual words between and within various rectangular ring regions,
and
{\bf (b)}~`block based ratio-occurrence' descriptor,
which describes local spatial information between visual words that could be stable.
For group image representation only features extracted from foreground pixels are used to construct visual words.

HPE~\cite{HPE_ICPR2010} considers multiple instances of each person to create a person signature.
The structural element (STEL) generative model approach~\cite{jojic2009stel} is employed for foreground detection.
The combination of a global (person level) HSV histogram and epitome regions of foreground pixels is then calculated,
where an image epitome \cite{jojic2003epitomic} is computed by collapsing the given image into a small collage of overlapped patches.
The patches contain the essence of textural, shape and appearance properties of the image.
Both the generic epitome (epitome mean) and local epitome (probability that a patch is in an epitome) are computed.

SDALF~\cite{SDALF_CVPR2010} considers multiple instances of each person.
Foreground features are used to model three complementary aspects of human appearance extracted from various body parts.
First, for each pedestrian image, axes of asymmetry and symmetry are found.
Then, complementary aspects of the person appearance
are detected on each part, and their features are extracted.
To select salient parts of a given pedestrian image,
the features are then weighted by exploiting perceptual principles of symmetry and asymmetry.

The above methods assume that classical Euclidean geometry is capable of providing meaningful solutions (distances and statistics)
for modelling and analysing images and videos, which might not be always correct~\cite{turaga2008statistical}.
Furthermore, they require separate parameter tuning for each dataset.

\section{Proposed approach}
\label{sec:Ourapproach}

Our goal is to automatically re-identify a given person
among a large set of candidates in diverse locations over various non-overlapping camera views.
The proposed method is comprised of three main stages:
{\bf (i)} feature extraction and generation of covariance descriptors,
{\bf (ii)} measurement of similarities on Riemannian manifolds via the Stein divergence,
and
{\bf (iii)} creation of similarity vectors and discriminative mapping for final classification.
Each of the stages is elucidated in more detail in the following subsections.

\subsection{Feature Extraction and Covariance Descriptors}
\label{Cov_Disc}

As per \cite{HPE_ICPR2010,SDALF_CVPR2010},
to reduce the effect of varying background,
foreground pixels are extracted from each given image of a person
via the STEL generative model approach~\cite{jojic2009stel}.
We note that it is also possible to use more advanced approaches, such as~\cite{Reddy_TCSVT_2013}.

Based on preliminary experiments,
for each each foreground pixel located at {\small $(x,y)$},
the following feature vector is calculated:

\vspace{-1ex}
\noindent
\begin{small}
\begin{equation}
  \Vec{f} = \left[ ~ x,~ y,~ \mathit{HSV}_{xy},~ \mathit{CIELAB}_{xy},~ \Lambda_{xy},~ \Theta_{xy}~ \right]^T
  \label{eq:features}
\end{equation}%
\end{small}%

\noindent
where
{\small $\mathit{HSV}_{xy}~\mbox{=}~[ H_{xy} , S_{xy} , \widehat{V}_{xy} ]$} are the colour values of the HSV channels,
employing histogram equalisation for channel~{\small $V$},
{\small $\mathit{CIELAB}_{xy}~\mbox{=}~\left[ L_{xy} , a_{xy} , b_{xy} \right]$} are the values of CIELAB colour space~\cite{ELAB_},
while
{\small $\Lambda_{xy}~\mbox{=}~[ \lambda_{xy}^{R}, \lambda_{xy}^{G}, \lambda_{xy}^{B} ]$}
and 
{\small $\Theta_{xy}~\mbox{=}~[ \theta_{xy}^{R}, \theta_{xy}^{G}, \theta_{xy}^{B} ]$}
indicate gradient magnitudes and orientations for each channel in RGB colour space.
We note that we have selected this relatively straightforward set of features as a starting point,
and that it is certainly possible to use other features.
However, a thorough evaluation of possible features is beyond the scope of this paper.

Given a set {\small $F = \left\{ \Vec{f}_i \right\}_{i=1}^{N}$} of extracted features,
with its mean represented by {\small $\Vec{\mu}$},
each image is represented as a covariance matrix:

\vspace{-1ex}
\noindent
\begin{small}
\begin{equation}
  \Mat{C} = \frac{1}{N-1}\sum\nolimits_{i=1}^{N}(\Vec{f}_i - \Vec{\mu} ){(\Vec{f}_i - \Vec{\mu})}^{T}
  \label{eq:Cov}
\end{equation}%
\end{small}%

Representing an image with a covariance matrix has several advantages~\cite{DPM_CVPR_2011}:
{\bf (i)} it is a low-dimensional (compact) representation that is independent of image size,
{\bf (ii)} the impact of noisy samples is reduced via the averaging during covariance computation,
and
{\bf (iii)} it is a straightforward method of fusing correlated features.

\subsection{Riemannian Manifolds and Stein Divergence}
\label{riemannian_geometry}

Covariance matrices belong to the group of symmetric positive definite (SPD) matrices,
which can be interpreted as points on Riemannian manifolds.
As such, the underlying distance and similarity functions might not be accurately defined in Euclidean spaces~\cite{Sanin_WACV_2013}.  

Efficiently handling Riemannian manifolds is non-trivial,
due largely to two main challenges~\cite{GDL_ECML_2011}: 
{\bf (i)}~as manifold curvature needs to be taken into account, defining divergence or distance functions on SPD matrices is not straightforward;
{\bf (ii)}~high computational requirements, even for basic operations such as distances.
For example, 
the Riemannian structure induced by considering the Affine Invariant Riemannian Metric (AIRM)
has been shown somewhat useful for analysing SPD matrices~\cite{hou2010efficient,raviv2011affine}.
For \mbox{\small $\Mat{A},\Mat{B} \in \mathcal{S}_{++}^d$},
where $\mathcal{S}_{++}^d$ is the space of positive definite matrices of size {\small $d \times d$},
AIRM is defined as:

\vspace{-3ex}
\noindent
\begin{small}
\begin{equation}
    \delta_R \left( \Mat{A}, \Mat{B} \right) \coloneqq
    \left\| \log \left( \Mat{B}^{-\frac{1}{2}} \Mat{A} \Mat{B}^{-\frac{1}{2}} \right) \right\|_F
    \label{eqn:AIRM}
\end{equation}
\end{small}

\vspace{-1ex}
\noindent
where {\small $\log(\cdot)$} is the principal matrix logarithm~\cite{Sra_JMLR_2012}.
However, AIRM is computationally demanding as it essentially needs eigen-decomposition of {\small $\Mat{A}$} and {\small $\Mat{B}$}.
Furthermore, the resulting structure has negative curvature
which prevents the use of conventional learning algorithms for classification purposes.

To simplify the handling of Riemannian manifolds,
they are often first embedded into higher dimensional Euclidean spaces,
such as tangent spaces~\cite{lui2011tangent,Porikli:2006:CT,Sanin_ICIP_2012,veeraraghavan2005matching}.
However, only distances between points to the tangent pole are equal to true geodesic distances,
meaning that distances between arbitrary points on tangent spaces may not represent the manifold accurately.

As an alternative to measuring distances on tangent spaces, 
in this work we use the recently introduced Stein divergence,
which is a version of the Bregman matrix divergence for SPD matrices~\cite{Sra_JMLR_2012}.
To measure dissimilarity between two SPD matrices {\small $\Mat{A}$} and {\small $\Mat{B}$},
the Bregman divergence is defined as~\cite{kulis2009low}:

\vspace{-1ex}
\noindent
\begin{small}
\begin{equation}
    D_{\phi}(\Mat{A},\Mat{B}) \triangleq  \phi( \Mat{A}) - \phi (\Mat{B}) -
    \left\langle \nabla_{\phi} (\Mat{B}) ,~ \Mat{A} - \Mat{B} \right\rangle
    \label{eqn:Bregman_Div}
\end{equation}
\end{small}

\vspace{-1ex}
\noindent
where {\small $\langle \Mat{A} , \Mat{B} \rangle = \tr \left( \Mat{A}^T \Mat{B} \right) $}
and {\small $\phi : \mathcal{S}_{++}^{d}\rightarrow \mathbb{R}$} is a real-valued,
strictly convex and differentiable function.
The divergence in \eqref{eqn:Bregman_Div} is asymmetric which is often undesirable.
The Jensen-Shannon symmetrisation of Bregman divergence is defined as~\cite{kulis2009low}:

\vspace{-1ex}
\begin{small}
\begin{equation}
    D_{\phi}^{JS}(\Mat{A},\Mat{B})
    \triangleq
    \frac{1}{2} D_{\phi}
    \left( \Mat{A},\frac{\Mat{A}+\Mat{B}}{2} \right)
    +
    \frac{1}{2} D_{\phi}
    \left( \Mat{B},\frac{\Mat{A}+\Mat{B}}{2} \right)
    \label{eqn:Jensen_Shannon_Div}
\end{equation}
\end{small}
\vspace{-1ex}

By selecting $\phi$ in \eqref{eqn:Jensen_Shannon_Div} to be {\small $- \log \left( \det \left( \Mat{A} \right) \right)$},
which is the barrier function of semi-definite cone~\cite{Sra_JMLR_2012},
we obtain the symmetric Stein divergence,
also known as the Jensen Bregman Log-Det divergence~\cite{cherian2011efficient}:

\vspace{-2ex}
\begin{small}
\begin{equation}
    J_{\phi }(\Mat{A},\Mat{B}) \triangleq  \log \left( \det \left( \frac{\Mat{A}+\Mat{B}}{2}\right) \right)
    - \frac{1}{2}  \log \left( \det \left( \Mat{A}\Mat{B} \right) \right)
    \label{eqn:Stein_Div}
\end{equation}%
\end{small}%

The symmetric Stein divergence is invariant under congruence transformations and inversion~\cite{cherian2011efficient}.
It is computationally less expensive than AIRM,
and is related to AIRM in several aspects
which establish a bound between the divergence and AIRM~\cite{cherian2011efficient}.

\subsection{Similarity Vectors and Discriminative Mapping}
\label{RDC}

For each query point (an SPD matrix) to be classified,
a similarity to each training class is obtained,
forming a similarity vector.
We obtain each similarity with the aid of the Stein divergence described in the preceding section.
The classification task on manifolds is hence converted into a task in the space of similarity vectors,
which can be tackled using learning methods devised for Euclidean spaces.

Given a training set of points on a Riemannian manifold,
{\small $\mathbb{X} =  \{ (\Mat{X}_1,y_1),~ (\Mat{X}_2,y_2),~ \ldots,~ (\Mat{X}_n,y_n)\}$},
where {\small $y_i \in \begin{Bmatrix} 1,2,\ldots,m \end{Bmatrix}$} is a class label,
and {\small $m$} is the number of classes,
we define the similarity between matrix $\Mat{X}_i$ and class $l$ as:

\vspace{-1ex}
\begin{small}
\begin{equation}
s_{i,l} = \frac{1}{N_l} {\sum\nolimits_{ j \neq i} J_{\phi }(\Mat{X}_i,\Mat{X}_j) \delta(y_j - l)}
\label{eq:clssim}
\end{equation}%
\end{small}%

\noindent
where {\small $\delta(\cdot)$} is the discrete Dirac function and

\vspace{-1ex}
\begin{small}
\begin{equation}
N_l = \left\{\begin{matrix}
n_l - 1~~~~\text{if}~~ y_i = l~~~~~
\\
n_l~~~~~~~~~~~\text{otherwise}~~~
\end{matrix}\right.
\end{equation}%
\end{small}%

\noindent
where {\small $n_l$} is the number of training matrices in class~{\small $l$}.
Using Eqn.~(\ref{eq:clssim}), the similarity between {\small $\Mat{X}_i$} and all classes is obtained,
where {\small $i \in \begin{Bmatrix} 1,2,\ldots,n \end{Bmatrix}$}.
Each matrix $\Mat{X}_i$ is hence represented by a similarity vector:

\vspace{-3ex}
\begin{small}
\begin{equation}
  \Mat{s}_i = \left[~ s_{i,1},~ s_{i,2},~ \ldots,~ s_{i,m}~ \right]^T
  \label{eq:simpat}
\end{equation}%
\end{small}%

Classification on Riemannian manifolds can now be reinterpreted as a learning task in {\small $\mathbb{R}^m$}.
Given the similarity vectors of training data,
{\small $\mathbb{S} = \{ (\Vec{s}_1,y_1), (\Vec{s}_2,y_2), \cdots, (\Vec{s}_n,y_n) \}$},
we seek a way to label a query matrix {\small $\Mat{X}_q$},
represented by a similarity vector
\mbox{\small $\Vec{s}_q = \left[ s_{q,1},~ s_{q,2},~ \ldots,~ s_{q,m}~ \right]^T$}.
As a starting point, we have chosen linear discriminant analysis~\cite{Bishop_2006},
where we find a mapping {\small $\Mat{W}^\ast$} 
that minimises the intra-class distances while simultaneously maximising inter-class distances:

\vspace{-3ex}
\begin{small}
\begin{equation}
\Mat{W}^\ast = \underset{\Mat{W}}{\operatorname{argmax}}~\operatorname{trace}
 \left\{
 \left[ \Mat{W} \Mat{S}_W \Mat{W}^T \right]^{-1} \left[\Mat{W} \Mat{S}_B \Mat{W}^T\right]
 \right\}
\label{eq:lda}
\end{equation}%
\end{small}%

\vspace{-1ex}
\noindent
where {\small $\Mat{S}_B$} and {\small $\Mat{S}_W$} are the between class and within class scatter matrices\cite{Bishop_2006}.
The query similarity vector {\small $\Vec{s}_q$} can then be mapped into the new space via:

\vspace{-2ex}
\begin{small}
\begin{equation}
   \Vec{x}_q = \Mat{{W}^\ast}^T \Vec{s}_q
   \label{eq:mappinQ}
\end{equation}
\end{small}

\vspace{-1ex}
We can now use a straightforward nearest neighbour classifier~\cite{Bishop_2006} to assign a class label to {\small $\Vec{x}_q$}.
We shall refer to this approach as Relational Divergence Classification (RDC).

\newpage
\section{Experiments and Discussion}
\label{sec:Experiments_RDC}

In this section we evaluate the proposed RDC approach by providing comparisons against several methods on two person re-identification datasets:
iLIDS~\cite{iLIDS_pub} and ETHZ~\cite{ETHZ_ICCV,schwartz2009learning}.
The  VIPeR dataset~\cite{VIPeR} was not used as it only has one image from each person in the gallery, and is hence not suitable for testing MvS approaches.
Each dataset covers various aspects and challenges of the person re-identification task.
The results are shown in terms of the Cumulative Matching Characteristic (CMC) curves,
where each CMC curve represents the expectation of finding the correct match in the top $n$ matches. 

In order to show the improvement caused by using similarity vectors in conjunction with linear discriminant analysis,
we also evaluate the performance of directly using the Stein divergence in conjunction with a nearest neighbour classifier
(ie.~direct classification on manifolds, without creating similarity vectors).
We refer to this approach as the {\it direct Stein} method.

\subsection{iLIDS Dataset}

The iLIDS dataset is a publicly available video dataset capturing real scenarios at an airport arrival hall under a multi-camera CCTV network.
From these videos a dataset of 479 images of 119 pedestrians was extracted
and the images were normalised to $128 \times 64$ pixels (height $\times$ width)~\cite{iLIDS_pub}.
The extracted images were chosen from non-overlapping cameras,
and are subject to illumination changes and occlusions~\cite{iLIDS_pub}.

\begin{figure}[!h]
\begin{minipage}{1\columnwidth}
  \begin{minipage}{1\columnwidth}
    \centering
    \includegraphics[width=1\columnwidth,keepaspectratio]{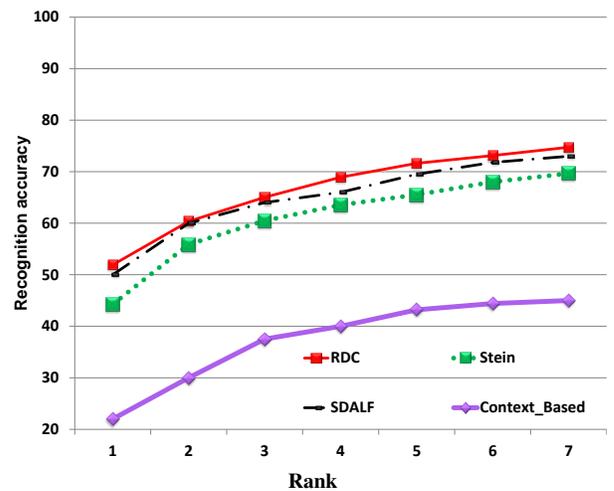}\\

    \vspace{-2ex}
    {\footnotesize\bf Rank}
    
    \vspace{-1ex}
    \caption
      {
      \small
      Performance on the iLIDS dataset~\cite{iLIDS_pub} for $N\mbox{=}3$,
      using the proposed RDC method,
      the direct Stein method,
      SDALF~\cite{SDALF_CVPR2010},
      context based method~\cite{iLIDS_pub}.
      HPE results for $N\mbox{=}3$ were not provided in~\cite{HPE_ICPR2010}.
      }
    \label{fig:iLIDS_B_3}
  \end{minipage}
  
\end{minipage}
\end{figure}

We randomly selected $N$ images for each person to build the gallery set, while the remaining images form the probe set. 
The whole procedure is repeated 10 times in order to estimate an average CMC curve.
We compared the performance of the proposed RDC approach against the direct Stein method,
as well as the algorithms described in Section~\ref{sec:Background}
(SDALF and Context based) for a commonly used setting of {\small $N$~=~$3$}.
The results, shown in Fig.~\ref{fig:iLIDS_B_3},
indicate that the proposed method generally outperforms the other techniques.
The results also show that the use of similarity vectors in conjunction with linear discriminant analysis
is preferable to directly using the Stein divergence.

\subsection{ETHZ Dataset}

The ETHZ dataset~\cite{ETHZ_ICCV,schwartz2009learning} was captured from a moving camera,
with the images of pedestrians containing occlusions and wide variations in appearance.
Sequence~1 contains 83 pedestrians (4857 images),
Sequence~2 contains 35 pedestrians (1936 images),
and Sequence 3 contains 28 pedestrians (1762 images).

We downsampled all the images to {\small $64 \times 32$} (height $\times$ width).
For each subject, the training set consisted of $N$ randomly selected images,
with the rest used for the test set.
The random selection of the training and testing data was repeated 10 times.

\begin{figure}[!t]
  \centering
  \includegraphics[width=0.9\columnwidth,keepaspectratio]{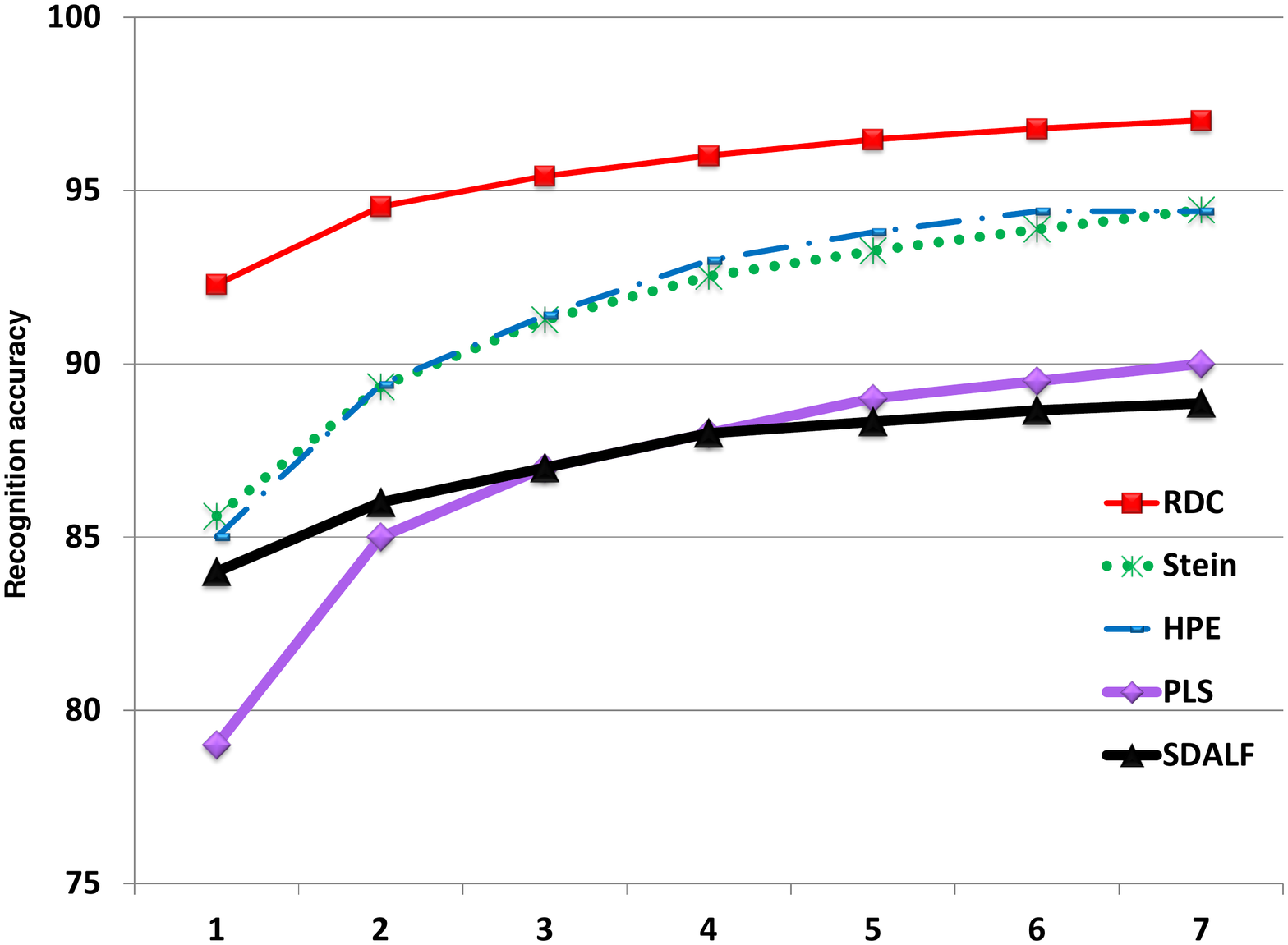}\\
  
  \includegraphics[width=0.9\columnwidth,keepaspectratio]{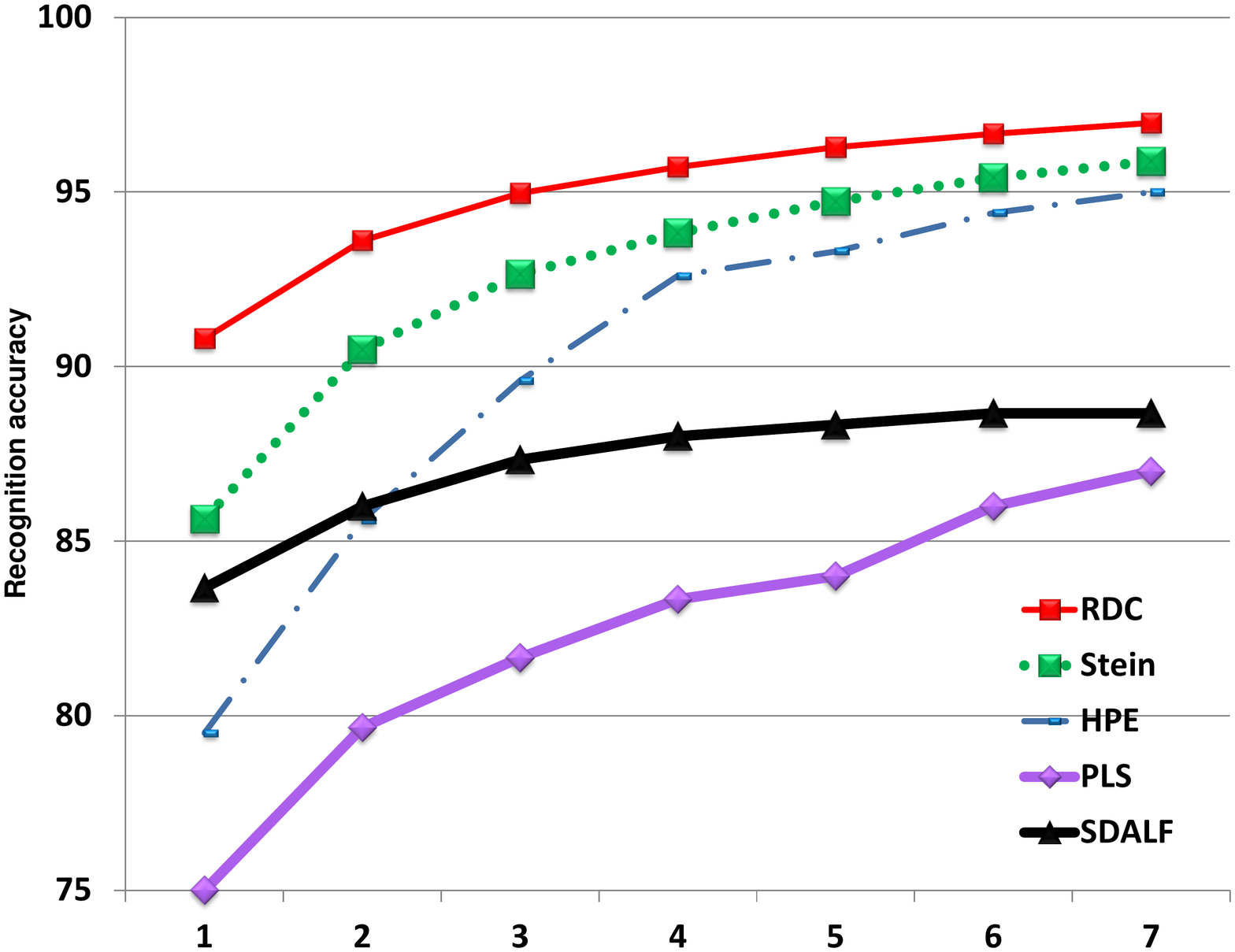}\\
  
  \includegraphics[width=0.9\columnwidth,keepaspectratio]{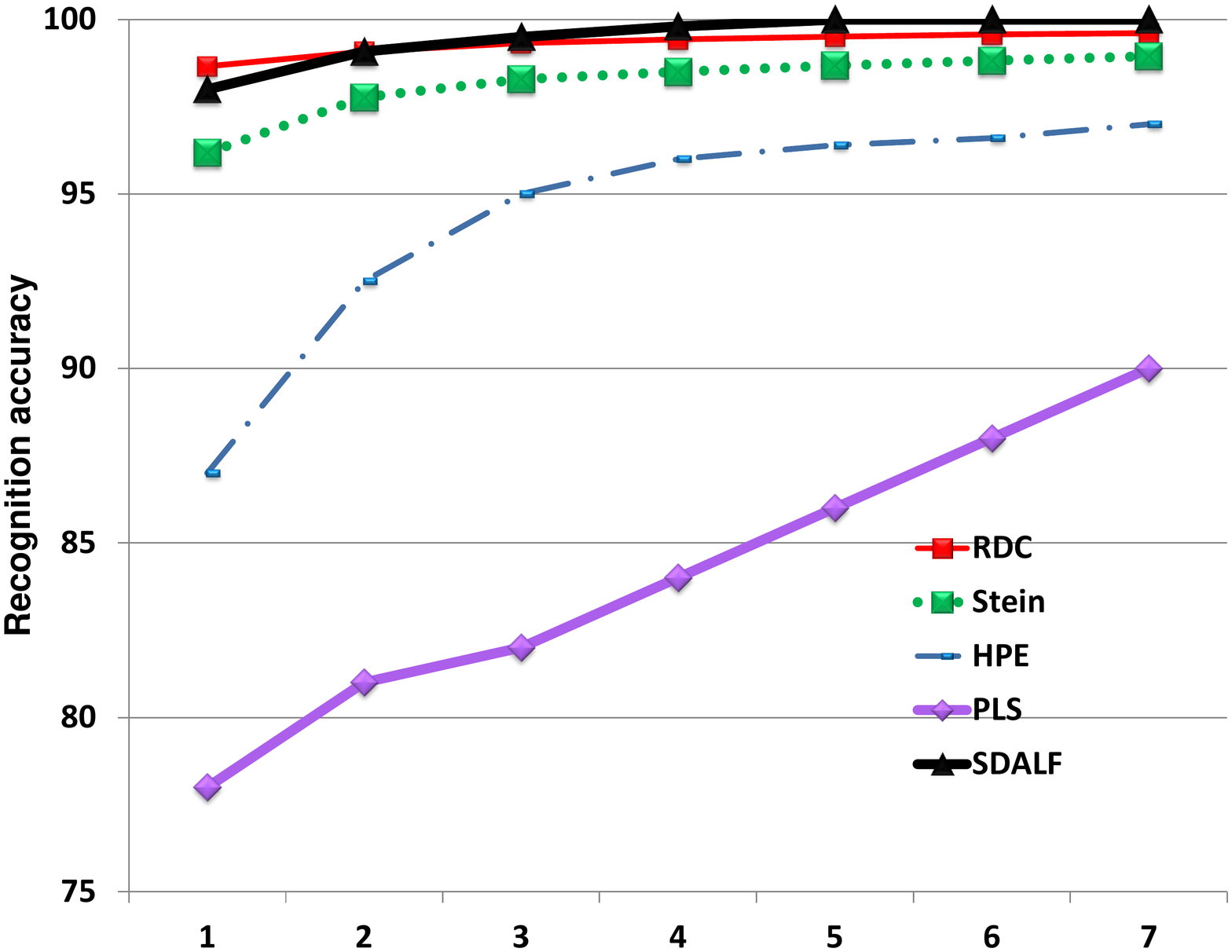}\\
  \vspace{-2ex}
  {\footnotesize\bf Rank}

  \vspace{-1ex}
  \caption
    {
    \small
    Performance on the ETHZ dataset~\cite{schwartz2009learning} for $N=10$, using Sequences 1 to 3 (top to bottom).
    Results are shown for the proposed RDC method,
    direct Stein method,
    HPE~\cite{HPE_ICPR2010},
    PLS~\cite{schwartz2009learning}
    and
    SDALF~\cite{SDALF_CVPR2010}.
    }
  \label{fig:ETHZ_B}
  
  ~
  
  ~
  
  ~
  
  ~
  
  ~
  
  ~
  
  ~
  
  ~
  
  ~
  
  ~
\end{figure}

Results were obtained for the commonly used setting of {\small $N$~=~$10$} and are shown in Fig.~\ref{fig:ETHZ_B}.
On sequences~1 and~2, the proposed RDC method considerably outperforms PLS, SDALF, HPE and the direct Stein method.
On sequence~3, RDC obtains performance on par with SDALF.

Note that the random selection used by the RDC approach to create the gallery is more challenging
and more realistic than the data selection strategy employed by SDALF and HPE on the same dataset~\cite{SDALF_CVPR2010,HPE_ICPR2010}.
SDALF and HPE both apply clustering beforehand on the original frames, 
and then select randomly one frame for each cluster to build their gallery set.
In this way they can ensure that their gallery set includes the keyframes to use for the multi-shot signature calculation.
In contrast, we haven't applied any clustering for the proposed RDC method in order to be closer to real life scenarios.

~
\section{Conclusion}
\label{sec:conclusions}
\vspace{-1ex}

We have proposed a novel appearance based person re-identification method comprised of:
{\bf (i)}~representing each image as a compact covariance matrix constructed from feature vectors extracted from foreground pixels,
{\bf (ii)}~treating covariance matrices as points on Riemannian manifolds,
{\bf (iii)}~representing each manifold point as a vector of similarities to class representers
with the aid of the recently introduced Stein divergence,
and
{\bf (iv)}~using a discriminative mapping of similarity vectors for final classification.
The use of similiarity vectors is in contrast to the traditional approach of analysing manifolds via embedding them into tangent spaces.
The latter might result in inaccurate modelling, as the structure of the manifolds is only partially taken into account~\cite{harandikernel,Harandi_ECCV_2012}.

Person re-identification experiments on the iLIDS~\cite{iLIDS_pub} and  ETHZ~\cite{ETHZ_ICCV,schwartz2009learning} datasets
show that the proposed approach outperforms several recent methods,
such as Histogram Plus Epitome~\cite{HPE_ICPR2010},
Partial Least Squares~\cite{schwartz2009learning},
and Symmetry-Driven Accumulation of Local Features~\cite{SDALF_CVPR2010}.

~

\section{Acknowledgements}

NICTA is funded by the Australian Government as represented by the {\it Department of Broadband, Communications and the Digital Economy},
as well as the Australian Research Council through the {\it ICT Centre of Excellence} program.

\clearpage
\balance
\bibliographystyle{ieee}
\bibliography{references}

\end{document}